\documentclass[10pt,twocolumn,letterpaper]{article}

\usepackage[applications]{wacv}      
\usepackage{amsmath,amsfonts}
\usepackage{algorithmic}
\usepackage{algorithm}
\usepackage{array}
\usepackage{tabularx}
\usepackage{textcomp}
\usepackage{stfloats}
\usepackage{url}
\usepackage{graphicx}
\usepackage{subcaption}
\usepackage{cite}

\usepackage{todonotes}
\usepackage{siunitx}  
\usepackage{ACIN_colors}
\usepackage{ACIN_macros}
\usepackage{multirow}
\usepackage{pifont}
\graphicspath{{graphics/}}
\usepackage{mathtools}

\newcommand{\optitrack}{OptiTrack}
\newcommand{\intelAero}{Intel\textregistered~Aero}

\DeclareSIUnit\event{Ev}
\DeclareSIUnit \pixel {px}
\DeclareSIUnit \pixres {p}
\DeclareSIUnit \fps {FPS}

\usepackage[pagebackref,breaklinks,colorlinks]{hyperref}
\usepackage[capitalize]{cleveref}
\crefname{section}{Sec.}{Secs.}
\Crefname{section}{Section}{Sections}
\Crefname{table}{Table}{Tables}
\crefname{table}{Tab.}{Tabs.}
\def\checkmark{\tikz\fill[scale=0.4](0,.35) -- (.25,0) -- (1,.7) -- (.25,.15) -- cycle;} 
\newcommand{\xmark}{\ding{55}}%
\usepackage{appendix}

\def\wacvPaperID{282} 
\def\confName{WACV}
\def\confYear{2024}
\newcommand\copyrighttext{%
  \footnotesize \textcopyright{ This work has been accepted to the IEEE/CVF Winter Conference on Applications of Computer Vision (WACV) 2024. Copyright may be transferred without notice, after which this version may no longer be accessible.}}
\newcommand\copyrightnotice{%
\begin{tikzpicture}[remember picture,overlay]
\node[anchor=south,yshift=10pt] at (current page.south) {\fbox{\parbox{\dimexpr\textwidth-\fboxsep-\fboxrule\relax}{\copyrighttext}}};
\end{tikzpicture}%
}
\begin{document}
\title{Real-time 6-DoF Pose Estimation by an Event-based Camera using Active LED Markers}
\author{Gerald Ebmer $^{1*}$ \and Adam Loch $^{1,2*}$ \and Minh Nhat Vu $^{1,2}$ \and Germain Haessig $^3$ \and Roberto Mecca $^2$  \and Markus Vincze $^1$\and Christian Hartl-Nesic $^1$ \and Andreas Kugi $^{1,2}$ 
}




\maketitle
\copyrightnotice

\def\thefootnote{*}\footnotetext{These authors contributed equally to this work.}
\def\thefootnote{$^1$}\footnotetext{Automation and Control Institute (ACIN), TU Wien, Vienna.}
\def\thefootnote{$^2$}\footnotetext{Austrian Institute of Technology (AIT), Vienna.}
\def\thefootnote{$^3$}\footnotetext{Prophesee GmbH, Paris.}
\begin{abstract}
Real-time applications for autonomous operations depend largely on fast and robust vision-based localization systems. 
Since image processing tasks require processing large amounts of data, the computational resources often limit the performance of other processes. 
To overcome this limitation, traditional marker-based localization systems are widely used since they are easy to integrate and achieve reliable accuracy. 
However, classical marker-based localization systems significantly depend on standard cameras with low frame rates, which often lack accuracy due to motion blur. 
In contrast, event-based cameras provide high temporal resolution and a high dynamic range, which can be utilized for fast localization tasks, even under challenging visual conditions. 
This paper proposes a simple but effective event-based pose estimation system using active LED markers (ALM) for fast and accurate pose estimation. 
The proposed algorithm is able to operate in real time with a latency below \SI{0.5}{\milli\second} while maintaining output rates of \SI{3}{\kilo \hertz}. 
Experimental results in static and dynamic scenarios are presented to demonstrate the performance of the proposed approach in terms of computational speed and absolute accuracy, using the OptiTrack system as the basis for measurement. 
Moreover, we demonstrate the feasibility of the proposed approach by deploying the hardware, i.e., the event-based camera and ALM, and the software in a real quadcopter application. Our project page is available at: \href{https://alecpose.github.io/almpose.github.io/}{almpose.github.io}
\end{abstract}


\section{Introduction}
\label{sec:intro}
Fast and reliable spatial localization is essential in a wide range of robotic applications. 
For example, in collaborative scenarios, the ability to accurately and rapidly estimate the pose of the end effector is a key component for achieving a safe, reliable, and robust execution of corresponding tasks. 
Vision-based methods \cite{kazerouni2022survey,zhao2019gslam,campos2021orb} are the most common approaches for obtaining the relative localization of objects within the line of sight. 
These methods achieve significantly better accuracy compared to other non-contact localization methods, \eg radio-based localization approaches ~\cite{wymeersch2020radio,heydariaan2018toward}. 
Vision-based approaches are, however, computationally expensive and typically require more than one sensor, \eg infrared-based systems~\cite{saeed2019state}.  
To reduce the computational overhead, classical markers \cite{kedilioglu2021arucoe,zhang2002visual} serving as easy-to-detect anchors are often integrated into vision-based systems. 
Since conventional RGB-D cameras are often used in these systems, the latency of detection cannot be reduced beyond the limit determined by the frame rate of the utilized cameras. 

\begin{figure}[t]
	\centering
        \def\svgwidth{0.9\columnwidth}
	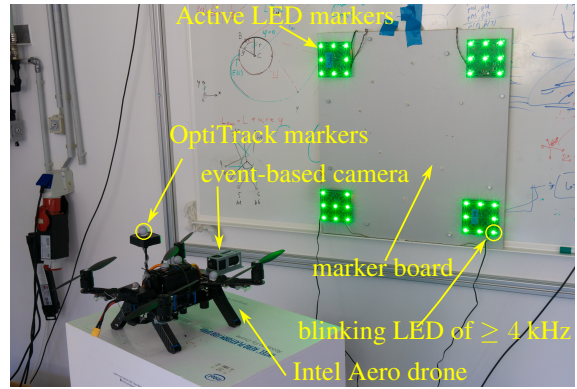
	\caption{Overview of the experimental setup. Active LED markers (ALM) are attached to a marker board. 
        An event-based camera mounted on a drone is used to estimate the pose of the marker board. 
        }
	\label{fig:overview}
\end{figure}

\textcolor{black}{
Event-based vision is an emerging field that has attracted much attention in recent years \cite{GaD22,nguyen2019real}. 
An event-based camera consists of an array of independent pixels measuring changes in luminosity $ L = \text{log}(I) $, based on the photocurrent $ I $ \cite{LiP08}. A change in the continuous luminosity signal
\begin{equation}
    \Delta L(\vec{u}_k,t_k) = L(\vec{u}_k,t_k) - L(\vec{u}_k,t_k - \Delta t_k) > p_k C
\end{equation}
triggers an event $ \vec{e}_k = (\vec{u}_k,t_k,p_k) $ at pixel location $ \vec{u}_k = (u_k,v_k) $
due to a temporal contrast threshold $\pm C$, $p_k \in \{+1, -1\}$ being its polarity and $\Delta t_k$ the time since the last event (at $\vec{u}_k$) occurred at $t_k$ \cite{GaD22}. The state-of-the-art event-based sensors can produce up to 1.2 Giga events per second (Geps)\cite{GaD22} with a microsecond range timestamp accuracy. 
}
\textcolor{black}{
Compared to frame-based cameras that deliver periodically dense (\ie full-frame) information, the event stream is sparse and contains information that relates only to changes in the scene. Additionally, event-based cameras have a high temporal resolution and a large dynamic range.
These features make them ideal for applications requiring fast and accurate detection. 
Starting from the early years of event-based vision development \cite{goal}, advantages given by those sensors for robotic applications are noticeable. 
Recent advancements in event-based sensor development \cite{GaD22}
have enabled them to compete with the precision of other localization methods \cite{ChY22} due to increased resolution and reduced noise. 
Avoiding accumulated event representations (\ie frames), markers can be tracked online utilizing the event-based camera's high temporal resolution of up to \SI{1}{\micro\second}.
}
\textcolor{black}{
An active LED marker (ALM) is a fixed geometric arrangement of individual LEDs, each unambiguously identifiable by its unique blinking frequency. By identifying the individual LEDs in the event stream and knowing the geometric arrangement of the LEDs, the pose of the ALM can be retrieved.
}

In this paper, we propose a fast and simple method employing an event-based camera together with ALM for simultaneous detection and tracking of the 6 degrees-of-freedom (DoF) pose of a rigid object in the 3D space. 
An overview of our proposed approach is depicted in Fig. \ref{fig:overview} with four ALMs attached to a marker board. 
The event-based camera is mounted on the drone, which is utilized to estimate the pose of the marker board with respect to the camera's base frame. 
To estimate the pose, the blinkings of the LEDs are logged with an event-based camera to identify the corresponding frequencies of each LED in the ALM. 
These blinking frequencies are utilized to identify each individual LED and match it with the known geometry of the ALM. 
With this mapping of the individual points on the camera's sensor plane and the known geometry of the ALM, the pose of the ALM can be computed by utilizing a Perspective-n-Point (PnP) algorithm. 
In the presented approach, by tuning the biases, i.e., parameters for tuning the analog front-end of the event-based camera, and using a priori knowledge about timing, the complexity of the ALM tracking can be simplified. 
This aids in reducing the tracking latency. 
During the tracking, the initial detection is continuously refined, resulting in subpixel resolution. 
Such an approach can still precisely estimate the pose even under fast rotational and linear motion.
The proposed approach was tested and verified extensively using an external infrared-based positioning system. 
Our contributions are listed in the following. 
\begin{itemize}
    \item We propose a fast event-based pose estimation system using ALM achieving a latency below  \SI{0.5}{\milli\second} while maintaining an output rate of 3 kHz. 
    \item 
    We analyze the proposed system in static and dynamic scenarios for several in-depth aspects, e.g., absolute accuracy, static noise, and latency. Translational errors of \SI{34.5}{\milli\m}$\pm$\SI{16}{\milli\m}  and \SI{0.74}{\degree}$\pm$\SI{0.15}{\degree} orientation errors at distances of \SI{2.1}{\m} to \SI{4.8}{\m} between the camera and the marker were achieved. Together with the fast computing speed, this proves that the proposed algorithm is promising for real-time applications. 
    
    \item We integrate the proposed system into a quadcopter application for the 6-DoF pose estimation task. For indoor experiments, the proposed system outperforms the ORB-SLAM algorithm. Furthermore, in outdoor experiments, the proposed system can simultaneously detect and track the ALM in very aggressive flights at velocities of up to \SI{10}{\m\per\second} and up to \SI{10}{m} away from the marker. 
\end{itemize}

The paper is organized as follows: Section \ref{sec:literature_review} presents the related work in the field of pose estimation with event-based cameras and active markers. 
Section \ref{sec:main_part} describes the proposed method for marker detection and tracking. 
In Section \ref{sec:experiment}, we present the experimental setup and results. 
Finally, we conclude the paper in Section \ref{sec:conclusion} with a summary of our contributions and suggestions for future work. 

\section{Related Work}
\label{sec:literature_review}
Visual localization systems show improved accuracy compared to systems based on other physical principles\cite{survey_indoor}, \cite{ChY22}. Fiducial marker-based systems \cite{survet_fiducial} constitute the most common choice for robotic applications. Due to the limited range and the dependence on the lighting conditions, some studies proposed LED-based solutions based on standard RGB cameras \cite{rgb_led}, infrared \cite{coded_infra}, or ultraviolet \cite{ultraviolet} spectrum. 
However, the latency cannot be reduced beyond the camera's frame rate. 

One of the first works in the direction of localization based on event-based sensors was the 2D localization method \cite{jorg}. The known shape (contours) was tracked, and the relative localization was determined by event-based vision. The high temporal resolution of the event-based sensors was used in \cite{MuH14} to localize an Unmanned Aerial Vehicle (UAV) during high-speed maneuvers. The pose information was retrieved using a black square as a known shape. In \cite{low_v_odometry}, a visual odometry method was proposed based on the feature tracking algorithm. In this direction, multiple methods were developed \cite{low_v_odometry}, \cite{cont}, which show a significant improvement compared to the RGB-based approach for high-speed applications.

The utilization of ALMs was proposed first in \cite{MuC11}, where the authors tracked the 2D position of the LED and used it as a feedback signal for the robot homing and a pan tilt system.
Later, the first method for pose estimation using ALMs was presented in \cite{CeS13}. 
Therein, ALMs were used to detect and estimate the position of a flying quadrocopter. 
LEDs were recognized and detected using event polarity changes in the event stream. 
In \cite{CeS13}, the authors used an accumulated event representation to decode the frequency and estimate the pose.
In \cite{crap}, a Gaussian mixture probability hypothesis density filter was proposed to localize the camera with respect to the active marker. Therein, online tracking was presented to increase the robustness and reliability of the pose estimation. 
The achieved results indicate a localization error lower than \SI{3}{\centi\metre} in scenarios where the camera was within \SI{1}{\metre} relative to the active marker.

Most recent works using ALMs propose the additional fusion of inertial measurements \cite{active_space}. 
The error in the predicted relative position is in the subcentimeter range. 
However, utilizing only the vision-based approach increases the error by the order of one magnitude. 
Compared to previous methods, the marker size is significantly larger. 
The LEDs are placed \SI{1}{\metre} apart. 
Current work in active marker-based solutions also focuses on the visual communication aspect of modulated light. 

Different from other approaches in the literature, our approach simplifies the complexity by tuning the biases and using a priori knowledge about timing.
This helps to reduce the tracking latency. 
To the best of the authors' knowledge, this work achieves the lowest latency compared to other methods in the literature. 
%


\section{Active Marker Tracking and Pose Estimation}
\label{sec:main_part}

\begin{figure*}
\centering
\def\svgwidth{0.9\textwidth}
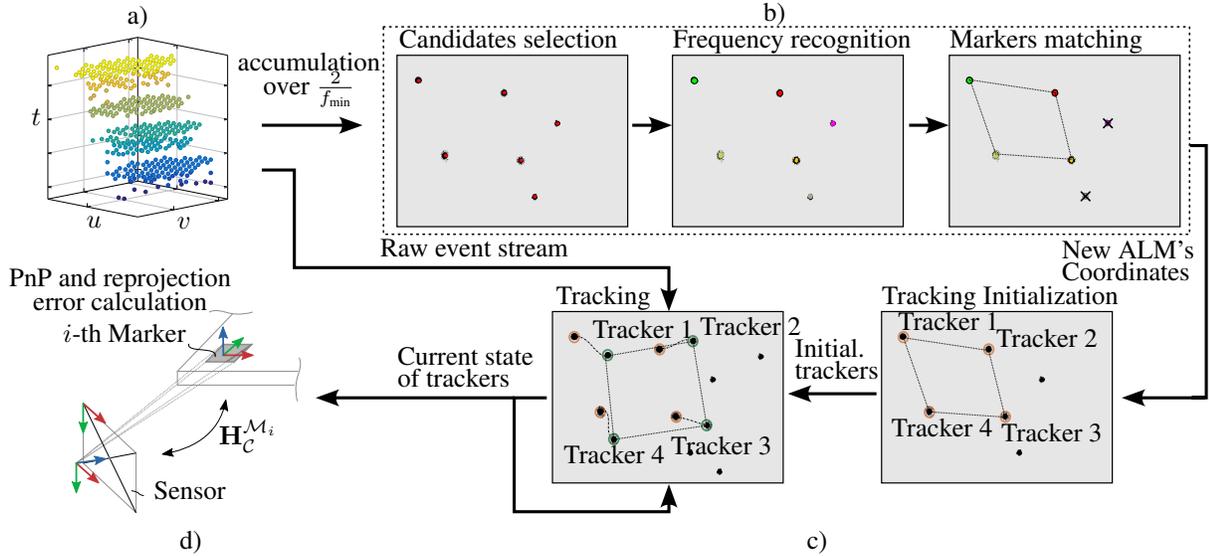
  \caption{Overview of the proposed approach.  
  Our pipeline consists of four asynchronous parts. 
  \emph{First}, illustrated in (a), to reduce the noise in the signal, biases are tuned to produce a single event per pixel on every blink. 
  The proposed detection algorithm accumulates the events over a short period of time (two times the period of the LED's minimal frequency $f_\text{min}$ to ensure at least two blinks are visible for every LED). 
  \emph{Second}, for the accumulated blinks, frequencies are recognized and assigned to the specified markers. 
  For every detected marker (b), trackers (c) are spawned for each individual LED. 
  \emph{Third}, using a simple tracking procedure, the LEDs are being tracked independently.
  \emph{Fourth}, to quantify tracking quality during runtime, the resulting solution (d) is used to compute the reprojection errors.}
  \label{fig:pipeline}
\end{figure*}

Using ALMs, periodic and dense signals can be generated as a projection of the LED on the camera's sensor plane.

To reduce the computational complexity and the required bandwidth, the biases of the sensor are tuned to generate a single event per pixel on every LED blink while suppressing all other background events to increase the signal-to-noise ratio, as presented in Figure \ref{fig:pipeline}. While \cite{CeS13} uses events of both polarities and relatively low frequencies ($1$-$2$kHz), the amount of noise can be reduced by using higher frequencies and disabling one polarity.

The ALM's structure is an arrangement of high-frequency blinking LEDs, where a unique frequency of the blinking pattern can individually recognize each LED (\eg different blinking frequencies). 
The arrangement of the LEDs has to be fixed and determined in the 3D coordinate space. 
However, it can also be arranged on a plane, as utilized in this work. Based on the 2D projection of the LEDs, knowing their 3D arrangement and camera intrinsics, the relative pose of the marker with respect to the sensor can be reconstructed using a Point-n-Perspective (PnP) algorithm \cite{pnp}. In this work, the \emph{IPPE} PnP algorithm is used \cite{CoB14}.

The proposed approach is divided into four parts, as illustrated in \cref{fig:pipeline}. To reduce the noise in the signal, the bias settings are tuned to produce a single event per pixel on every blink of an LED. Next, events are accumulated over the time $\frac{2}{f_\text{min}}$, which is two times the period of the LED's minimal frequency $f_\text{min}$ (typ. $f_\text{min} \approx \SI{2}{\kilo\hertz}$).
For those accumulated event clusters, frequencies are recognized. These frequencies are used to identify newly appearing ALMs. For each of the ALM's LEDs, trackers are spawned that keep track of the LEDs' center points based on single events. The tracking of the LEDs is independent of the detection loop.
The pose of the ALM is estimated utilizing the trackers of the ALM. The accuracy of the pose estimation can be obtained using the reprojection error. When an ALM leaves the field of view or the reprojection error exceeds the defined maximal value, the corresponding trackers are deleted. If the ALM enters the field of view again, the detection algorithm respawns it. Such an approach reduces the latency and increases the accuracy of the solution.

\subsection{Detection}
\label{sec:marker_detection}
For the detection of an ALM, the geometrical arrangement and the blinking frequency of the ALM LEDs, have to be provided in advance. 

The range of possible frequencies for the LEDs is wide: from tests conducted, frequencies higher than 4kHz and lower than 40kHz work best. To detect lower frequencies, biases have to be adapted to maximize the signal-to-noise ratio. As the timestamp is quantized, it is advisable to use LED frequencies with an integer microsecond period. 

Due to the limited noise, detection can be simplified by using only single types of events. In \cite{active_space} and \cite{CeS13}, the detections rely on the transitions between event polarities. Instead, in the proposed approach, we use the timing information between consecutive events generated by a single pixel. 

For detection, an event frame generated over the period $T_d$ is used, where $T_d$ has to be larger than $\frac{2}{f_\text{min}}$ to ensure that at least two blinks are visible for every LED. Candidates for the blinking LEDs can be retrieved by selecting the connected regions where more than $T_d f_\text{min}$ events per pixel are generated. Each region with an area larger than a defined minimal area is selected as a potential candidate. Due to the short accumulation period, even under fast motion, the LEDs' center points can be calculated by computation of the center of mass on a 2D plane. The error introduced by the relative movement of the LED is refined by the tracking procedure.

For the frequency estimation of the LEDs, a histogram of the time differences between events in a given area is used. In the case of frequencies with an integer microsecond period, the histogram has a pronounced peak, while for other frequencies, the histogram follows a wider Gaussian distribution. The frequency estimation follows the procedure proposed in \cite{CeS13}. 

\subsection{Tracking}
\label{sec:marker_tracking}
While detection relies on an accumulated representation of the events, the tracking can be performed online to reduce latency. Using the initial guess from the detection of the ALM's LED center points, trackers are spawned for every LED. The $i$-th tracker is characterized by its frequency $f_i$, center point $\vec{c}_i=[x_i, y_i]$, and radius $r_i$. In comparison to the assumptions of \cite{multi_kernel} and \cite{part_based}, the distribution of the generated events (within one blink) follows a spatially uniform distribution and hence, produces a dense event stream in this region. This allows us to simplify the tracking algorithm while maintaining precise tracking with sub-pixel accuracy.

For every LED's blink, the tracker's center of mass $\bar{\vec{c}}_{i}$ is calculated using all events within its current radius $r_i$. The update term 
\begin{equation}
    \bar{\vec{c}}_{i} = \tilde{\beta} \vec{u}_{k} + (1-\tilde{\beta}) \bar{\vec{c}}_{i} 
\end{equation}
introduces low-pass filtering, where every new event $\Vec{u}_k$ updates the current solution directly with an update factor $\tilde{\beta}$ of typically 0.02. The radius $r_i$ is updated every $N$ events and set to twice the average distance of the events from the center point of the tracker.

\subsection{Pose Estimation}
\label{Pose estimation and verification}
The 6-DoF pose is estimated asynchronously, using the current center points of the ALM's trackers. To increase the update rate of the algorithm, a PnP algorithm is started whenever the previous iteration is done. Due to the simplicity of the tracking, the PnP calculation is decisive in terms of latency and output rate.

To ensure stability and detect tracking failures, the reprojection error is computed and compared to the tracker's center points. When the reprojection error of one tracker exceeds the mean distance of the events from the center point, a tracking lost signal is generated, and tracking is stopped. It is reinitialized with the first new detection of a given marker.

\section{Experiments}
\label{sec:experiment}

For the experimental setup, the EVK4 HD evaluation kit from Prophesee is used. It includes the event-based vision sensor IMX636ES providing HD resolution ($1280\times 720$ pixels) and the Soyo SFA0820-5M lens. The ALM consists of printed circuit boards with 8 LEDs arranged in a square of \SI{9}{\centi \m} side length. Each ALM has a base frequency (first LED), and the remaining frequencies are selected to match integer microsecond period times. For the experiments, four markers are arranged in a square on a marker board with a side length of \SI{59}{\centi \m}. The 8 outermost LEDs were chosen to create a single marker.
The event stream is processed on a Desktop PC (Ubuntu 20.04, Intel i9-12900K, 32GB RAM) and on an Intel Aero Compute Board. 
As ground truth, the commercial infrared 3D tracking system \optitrack{} is used. The \optitrack{} recordings are triggered with the same trigger signal as the event-based camera via its external trigger input.

\subsection{Bias Adjustment}
\label{sec:bias}

\begin{figure}
  \centering
  \begin{subfigure}[]{0.49\columnwidth}
    \centering
    \includegraphics[]{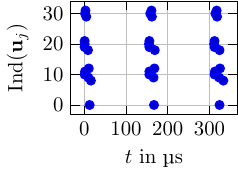}
    \caption{Adjusted biases}
    \label{fig:bias_good}
  \end{subfigure}
  \hfill
  \begin{subfigure}[]{0.49\columnwidth}
    \centering
    \includegraphics[]{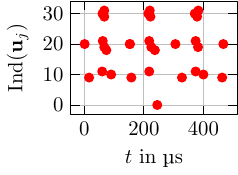}
    \caption{Default biases}
    \label{fig:bias_bad}
  \end{subfigure}
  \begin{subfigure}[]{\columnwidth}
    \centering
    \includegraphics[]{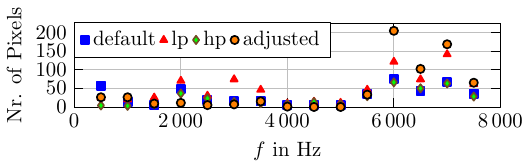}
    \caption{\textcolor{black}{Frequency histogram}}
    \label{fig:bias_hist}
  \end{subfigure}
  \caption{Visualization of bias adjustment. The vertical axis in \subref{fig:bias_good} and \subref{fig:bias_bad} represents the flattened indices of the pixels in the region of interest (ROI) around an LED light. Plot \subref{fig:bias_good} and \subref{fig:bias_bad} demonstrate the event distribution with optimal and default bias settings, respectively. 
  \textcolor{black}{
  In the frequency histogram \subref{fig:bias_hist}, the effect of the low-pass (lp) and high-pass (hp) settings are illustrated beside the default and adjusted bias settings.
  }
  } 
  \label{fig:bias}
\end{figure}


The bias adjustment of the event camera is essential for the proposed system's performance. The IMX636ES sensor biases \cite{metavisionbiases} allow control over analog pixel gate thresholds to achieve the desired sensor response. The adjustment goal is to minimize the number of activated pixels between two LED blinks, as shown in Figure \cref{fig:bias}, thereby reducing processing complexity. The proposed method employs a single event polarity for simplicity.

By adjusting the refractory period setting, a pixel should be rendered insensitive to subsequent changes in LED brightness. An optimal value during adjustments should filter out all events between two consecutive LED-triggered events, as depicted in \cref{fig:bias_good}. By utilizing high-pass and low-pass filter setups, the number of environment-generated events (excluding those by LEDs) can be limited to prevent sensor overflow and maintain manageable event blob density.
\textcolor{black}{
The histogram of frequencies at which events occur over an accumulation time of \SI{10}{\milli\second} is depicted in Figure \ref{fig:bias_hist}. It illustrates the effect of the low-pass (lp) and high-pass (hp) bias settings acting as a band-pass filter for the LED frequencies of the ALM. 
\textcolor{black}{Please note that the bias values may affect each other \cite{metavisionbiases}. 
This causes the high event count, with the adjusted bias settings, at \SI{6}{\kilo \hertz} in Fig. \ref{fig:bias_hist}, where the value exceeds low-pass and high-pass settings. }
}

A detailed explanation of the sensor biases is provided in \cite{metavisionbiases} and the description of the bias adjustment procedure is detailed in the supplementary document. 

\subsection{Absolute Accuracy}
\label{sec:abs_accuracy_estimation}

To evaluate the absolute accuracy, the pose estimation of the ALMs and the marker board are compared with the synchronized measurements of the \optitrack{} system (ground truth). 
For this experiment, the marker board is placed statically in the scene. The camera moves from close to far, covering the working distance of the setup, which is limited by the \optitrack{} setup. The kinematic relations of the experimental setup are described in detail in the supplementary document.

\begin{figure}
  \centering
  \includegraphics[width=\columnwidth]{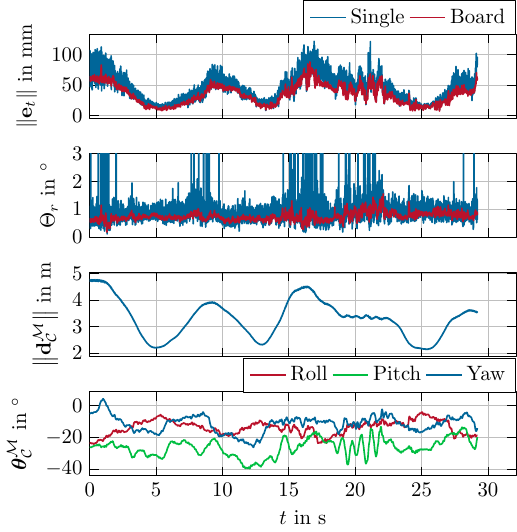}
  \caption{
  Analysis of the orientation and position error with reference distances and orientations.
  }
  \label{fig:abs_acc}
\end{figure}


The magnitude of the absolute position error $\vec{e}_{t}$ and the orientation error $\Theta_{r}$ with the distance between the marker and the camera $\norm{{\vec{d}}_\mathcal{C}^\mathcal{M}}$, ranging from $\SI{2.1}{\m}$ to $\SI{4.8}{\m}$,  as well as the orientation $\vec{\Theta}_\mathcal{C}^{\mathcal{M}}$ of the marker with respect to the camera, is depicted in \cref{fig:abs_acc}. 
Moreover, \cref{fig:abs_acc} illustrates the difference of using a single ALM with a side length of \SI{9}{\centi \metre} for pose estimation compared to a larger marker board with a side length of \SI{59}{\centi \metre}. From the plot of the position error $\norm{\vec{e}_t}$ in \cref{fig:abs_acc} it can be seen that the marker board has less noise but a comparable error magnitude. The second plot displaying the orientation error $\Theta_r$ shows significant spikes for the single ALM. This indicates flips in the estimated pose, especially for medium to far distances. Hence, the usage of a marker board with increased side length is beneficial for accurate orientation estimations. 
The plot of the marker orientation $\vec{\Theta}_\mathcal{C}^{\mathcal{M}}$ in \cref{fig:abs_acc} shows fast orientation changes beginning at \SI{20}{\second}. This demonstrates the ability of the proposed method to accurately estimate pose information even in highly dynamic scenes. 

\begin{figure}
  \centering
  \includegraphics[width=\columnwidth]{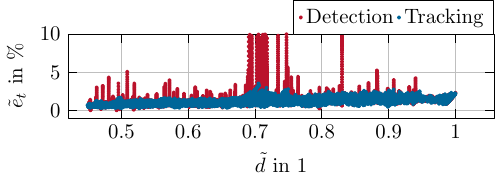}
  \caption{
  Normalized relative position error ${\tilde{e}_t}$.
  }
  \label{fig:framed_tracking_comparison}
\end{figure}

In order to compare the performance between the detection-based pose estimation (\cref{sec:marker_detection}) and the tracking-based pose estimation (\cref{sec:marker_tracking}), the relative position error $ {\tilde{e}_t} =  \frac{\norm{\vec{e}_t}}{\norm{{\vec{d}}_\mathcal{C}^\mathcal{M}}}$ is displayed in \cref{fig:framed_tracking_comparison} as a function of the normalized distance $ \tilde{d} = \frac{\norm{\vec{d}}_\mathcal{C}^\mathcal{M}}{\max \norm{\vec{d}}_\mathcal{C}^\mathcal{M}}$.
The results for the tracking-based approach indicate a better consistency, \ie less noise, and altogether lower error numbers in the \SI{1}{\percent} to \SI{2}{\percent} range.
The expected linear increase of the position error with the distance can also be inferred from \cref{fig:framed_tracking_comparison}. 
The statistical values of the data illustrated in \cref{fig:framed_tracking_comparison} are summarized in \cref{tab:abs_err_stat}. The maximum position error of \SI{87.8}{\milli\m} at a distance of \SI{4.8}{\m} and the maximum orientation error of \SI{1.55}{\degree} indicate the excellent performance of the tracking-based approach. The standard deviation of the position and orientation error of \SI{16.2}{\milli \m} and \SI{0.146}{\degree}, respectively, show the robustness of our method. In Table \ref{tab:comparison_sota}, we contextualize our results within different types of positioning systems.

\begin{table}[h]
\centering
\caption{Statistical values of the absolute accuracy measurements. }
\label{tab:abs_err_stat}
\footnotesize
\begin{tabular}{l c c c c }
            & \multicolumn{2}{c}{Tracking}               & \multicolumn{2}{c}{Detection} \\
            & $\norm{\vec{e}_t}$   & $\Theta_r$          & $\norm{\vec{e}_t}$   & $\Theta_r$  \\ \hline
Mean        & \SI{34.5}{\milli \m} & \SI{0.738}{\degree} & \SI{64.9}{\milli \m} & \SI{1.55}{\degree}   \\
Std. Dev.   & \SI{16.2}{\milli \m} & \SI{0.146}{\degree} & \SI{121}{\milli \m} & \SI{5.12}{\degree}   \\
Maximum     & \SI{87.8}{\milli \m}   & \SI{1.55}{\degree}  & \SI{1.233}{ \m}   & \SI{71.9}{\degree}   \\ \hline
\end{tabular}

\end{table}

\subsection{Static Noise}
In \cref{tab:statis_noise}, the noise floor of the proposed method is characterized by different distances between the camera and the marker board. The low standard deviation values indicate the stability of the pose estimation. This data can be utilized to tune Bayesian filters (\eg Kalman filters).

\begin{table}[h]
\centering
\caption{Statistics of the noise in static scenes. }
\label{tab:statis_noise}
\footnotesize
\begin{tabular}{l c c c}
 $\norm{{\vec{d}}_\mathcal{C}^\mathcal{M}}$  & Std. Dev. & Maximum \\ \hline
 \SI{6}{\m}   & \SI{1.4}{\milli \m} & \SI{5.5}{\milli \m}  \\
 \SI{4}{\m}   & \SI{0.68}{\milli \m}  & \SI{2.95}{\milli \m}   \\
 \SI{2}{\m}   & \SI{0.25}{\milli \m} & \SI{2.17}{\milli \m}   \\ \hline
\end{tabular}

\end{table}

\subsection{Latency Measurement and Output Rate}
\label{sec:latency_meas}

To determine the latency of the proposed system, the execution priority was elevated. Additionally, the visualization, as well as background tasks of the operating system, were disabled. This avoids unintentional interrupts and stalls during the execution of the pose estimation. 

The latency and output rate values are listed in \cref{tab:latency}. The output rate of the tracking-based approach outperforms the detection-based method while achieving comparable latency results. As shown in \cref{tab:latency}, the proposed method is capable of running even on an embedded PC of a drone. While maintaining real-time performance, we can notice a reduced output rate (limited by the PnP computation time) as well as an increased average delay (limited by a number of concurrent threads) compared to a desktop PC. Latency is measured using a precise synchronization trigger signal and is equal to the time difference between the trigger and the time when pose estimation for this timestamp is available.
Our proposed method achieves lower mean latency combined with low standard deviation compared to the state of the art. 
A large part of the resulting latency is due to communication overhead. 



\begin{table}[h]
\centering
\caption{Latency and output rates using a Desktop PC (PC) and Intel Aero Compute Board (Drone).}
\footnotesize
\begin{tabular}{l l c c c c }
            &          & \multicolumn{2}{c}{Tracking}               & \multicolumn{2}{c}{Detection} \\
            &          & latency   & rate          & latency   & rate  \\ \hline
\multirow{2}{*}{PC} & Mean        & \SI{354}{\micro\second} & \SI{3.81}{\kilo\hertz} & \SI{699}{\micro\second} & \SI{670}{\hertz}   \\
            & SD   & \SI{92}{\micro\second} & \SI{64}{\hertz} & \SI{35}{\micro\second} & \SI{288}{\hertz}   \\
            \hline 
\multirow{2}{*}{Drone} & Mean        & \SI{1232}{\micro \second} & \SI{1.32}{\kilo \hertz} & \SI{1953}{\micro \second} & \SI{223}{ \hertz}   \\
            & SD   & \SI{194}{\micro \second} & \SI{140}{ \hertz} & \SI{240}{\micro \second} & \SI{94}{ \hertz}   \\
\hline
\end{tabular}
\label{tab:latency}
\end{table}

\begin{table*}
    
\caption{\textcolor{black}
    {Comparison of different marker-based approaches and visual odometry/SLAM methods to contextualize the performance of the system. All the results are taken from corresponding papers. *Possible types E-Events, F-Frames, I-IMU. \dag Absolute Trajectory Error (RMS) reported in \cite{hidalgo2022event} using "kitchen" sequence. \ddag Sequences used for evaluation include only slow motion. 
    }
}
\centering
\def\svgwidth{0.9\textwidth}
\footnotesize
\begin{tabular}{c|c|c|c|c|c|c|c|c}
\multicolumn{1}{c|}{Method} & Input* & Rate/FPS & Range       & Markers &  \begin{tabular}[c]{@{}c@{}}Positioning \\ error\end{tabular} & Resolution & \begin{tabular}[c]{@{}c@{}}Dynamic \\ motion\end{tabular} & \begin{tabular}[c]{@{}c@{}}Marker\\ size\end{tabular} \\ \hline
Ours                        & E    & \SI{3.8}{\kilo \hertz} & up to \SI{10}{\m}  & Active  & \begin{tabular}[c]{@{}c@{}} \SI{1.89}{\percent} \\ \SI{2.11}{\percent} \end{tabular}    & \SI{720}{\pixres}       & \checkmark                                                         & \begin{tabular}[c]{@{}c@{}}\SI{59}{\times}\SI{59}{\centi\m} \\ \SI{9}{\times}\SI{9}{\centi\m}\end{tabular}                                        \\
\begin{tabular}[c]{@{}c@{}}Censi et al.\\ 2013\cite{CeS13}\end{tabular}                   & E    & \SI{250}{\hertz}   & -           & Active  & \SI{8.9}{\centi\m}            & \SI{128}{\times} \SI{128}{}    & \checkmark                                                         &           \SI{20}{\times}\SI{20}{\centi\m}                                          \\
\begin{tabular}[c]{@{}c@{}}Salah et al.\\ 2022\cite{active_space}\end{tabular}                 & E+I  & \SI{200}{\hertz}   & up to \SI{7}{\m}   & Active  & \SI{0.074}{\percent}          & \SI{480}{\pixres}       & \xmark \ddag  & \SI{3}{\times}\SI{3}{\times}\SI{3}{\m}                                                \\
\begin{tabular}[c]{@{}c@{}}Chen et al.\\ 2020\cite{crap}\end{tabular}                       & E    & -        & \SI{1}{\m} (tests) & Active  & \SI{3}{\percent}              & \SI{346}{\times} \SI{240}{}    & \xmark     & \SI{40}{\times}\SI{30}{\centi\m}                                              \\
STag \cite{benligiray2019stag}                      & F    & \SI{56}{\fps}   & up to \SI{3}{\m}    & Passive & \SI{1.32}{\percent}   & \SI{720}{\pixres}       & \xmark                                                         & \SI{15}{\times}\SI{15}{\centi\m}                                              \\
ORB-SLAM2 \cite{mur2017orb}                   & F    & -        & -           & SLAM    & \SI{13.0}{\centi\m}\dag         & \SI{480}{\pixres}       & \checkmark                                                         &   -                                                    \\
EDS  \cite{hidalgo2022event}                       & E+F  & -        & -           & VO      & \SI{9.6}{\centi\m}\dag           & \SI{480}{\pixres}       & \checkmark                                                         &  -                                                    
\end{tabular}
\label{tab:comparison_sota}
\end{table*}

\subsection{Application to 6-DoF position estimation for a quadcopter}
\label{sec:drone_control}

In this section, indoor and aggressive outdoor flights are considered for the 6-DoF position estimation of a quadcopter. 
\subsubsection{Indoor flight experiments}
Compared to stationary robots, e.g. articulated manipulators \cite{VU2023102970,beck2022singularity}, Cartesian robots \cite{vu2020fast,vu2022fast,vu2022sampling,o2003gantry}, which are equipped with high-precision encoders to monitor their state, flying robots \cite{zimmermann2022two} mainly rely on IMUs, barometers, and vision-based systems to estimate their state. 
While the pose estimation module equipped with only IMUs and barometers often suffers from the problem of drift, vision-based systems ensure a more reliable measurement. 
In this experiment, the \intelAero{} Ready to Fly (RTF) Drone, shown in Fig. \ref{fig:overview}, is employed as it offers enough computational power for on-board processing. 

\begin{figure}
    \centering
    \includegraphics[width = 1\columnwidth]{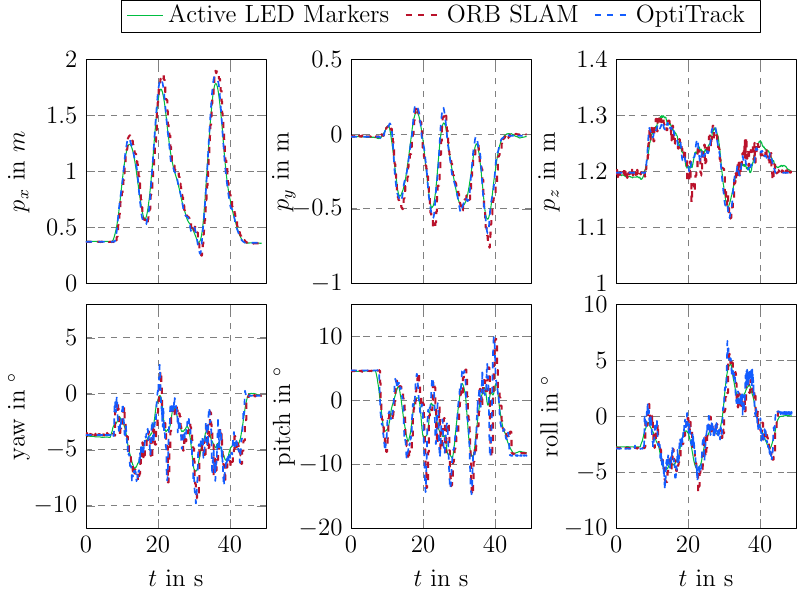}
    \caption{Time evolution of the drone trajectory. 
    }
    \label{fig: trial 2}%
\end{figure}

\begin{figure}
    \centering
    \includegraphics[width = 1\columnwidth]{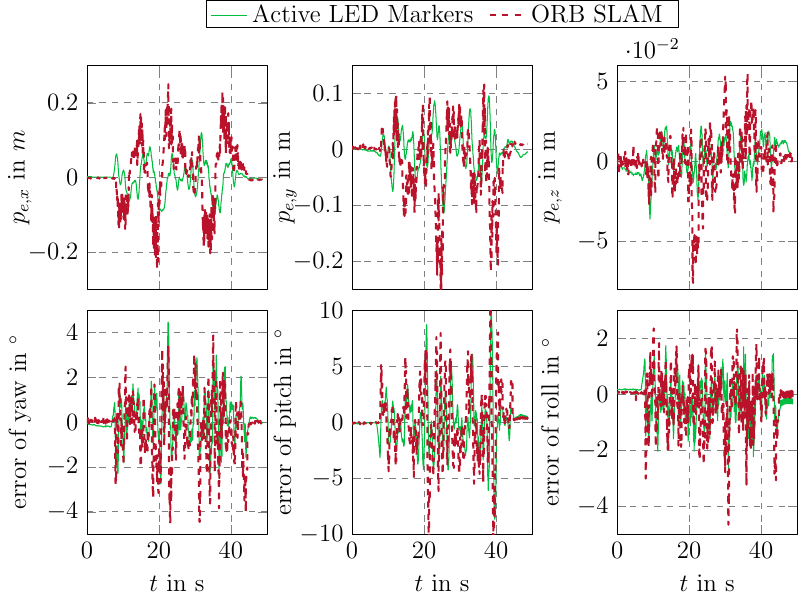}
    \caption{The error plots of the corresponding estimates, depicted in Fig. \ref{fig: trial 2} with respect to the ground-truth measurements. 
    }%
    \label{fig: trial 2 error}%
\end{figure}

For absolute drift-free pose information, the  ORB-SLAM2 \cite{mur2017orb} algorithm is utilized. 
The ORB-SLAM2 was chosen for its impressive performance and open-source implementation. 
Note that ORB-SLAM3 \cite{campos2021orb}, as the successor of ORB-SLAM2, is a more robust version compared to ORB-SLAM2. However, the accuracy of these approaches on stereo and RGB-D cameras is still comparable since the key concepts of the estimation module and the relocalization method, remain unchanged. 
We are aware that comparing the proposed system with other SLAM algorithms, e.g., feature-based SLAM \cite{mur2017orb} and event-based SLAM \cite{chamorro2022event}, may not be fair because the key concept is different. However, this comparison could provide a qualitative guide for choosing the right method in a given situation and contextualize results as shown in Table \ref{tab:comparison_sota}. Similar to previous subsections, the \optitrack{} serves as the source of ground truth. 
For the position errors, the metric for comparison is the difference between a ground-truth position and the estimated position as 
    $\vec{p}_e = [p_{e,x}\:p_{e,y}\:p_{e,z}]^\mathrm{T}= \vec{p}_g - \hat{\vec{p}} \Comma$
where the hat ($\:\hat{}\:$) indicates quantities estimated by the ORB-SLAM2 algorithm \cite{mur2017orb} or with the event-based marker, respectively, the subindex $(\cdot)_g$ stands for the three-dimensional ground-truth quantities, and the subindex $(\cdot)_e$ for the resulting three-dimensional errors.
The orientation errors are represented as Euler angles.
The difference between the ground-truth quaternion and the estimated quaternion is defined as 
    $\vec{q}_e = \hat{\vec{q}}^{-1} \otimes \vec{q}_g$ \Comma
with $\otimes$ as the quaternion product.
Subsequently, this error quaternion can be transformed into an equivalent representation using three angles, i.e., roll, pitch, and yaw. 
In this experiment, the quadcopter is moved aggressively in a zigzag pattern in a space of \SI{1.8}{\meter} in the $x$-direction, \SI{0.6}{\meter} in  $y$-direction, and \SI{0.4}{\meter} in  $z$-direction for about \SI{50}{\second}. 
The position estimates and the resulting errors for this case are illustrated in Fig. \ref{fig: trial 2} and Fig. \ref{fig: trial 2 error}. 
The errors obtained from the proposed algorithm in the $x$-, $y$- and $z$-direction are bounded within $\pm \SI{0.06}{\meter}$, $\pm \SI{0.08}{\meter}$, and $\pm \SI{0.02}{\meter}$, respectively, while larger errors result from the ORB-SLAM2, i.e., $p_{e,x}=\pm\SI{0.2}{\meter}$, $p_{e,y}\in [-0.2,0.1]^\mathrm{T}\:\SI{}{\meter}$, and $p_{e,z}=\pm \SI{0.05}{\meter}$. 
The orientation errors achieved with the two methods are similar, shown at the bottom of Fig. \ref{fig: trial 2 error}. However, the orientation errors measured by the proposed system are slightly better since the spikes in ORB-SLAM2 are larger. More experiments can be found in the supplementary material. 



\subsubsection{Outdoor Flight Experiments}
\begin{figure}
    \centering 
    \hfill
    \begin{subfigure}{\columnwidth}
        \begin{minipage}[t]{0.33\columnwidth}
            \vspace{5pt} 
            \centering
            \def\svgwidth{\textwidth}
\begingroup%
  \makeatletter%
  \providecommand\color[2][]{%
    \errmessage{(Inkscape) Color is used for the text in Inkscape, but the package 'color.sty' is not loaded}%
    \renewcommand\color[2][]{}%
  }%
  \providecommand\transparent[1]{%
    \errmessage{(Inkscape) Transparency is used (non-zero) for the text in Inkscape, but the package 'transparent.sty' is not loaded}%
    \renewcommand\transparent[1]{}%
  }%
  \providecommand\rotatebox[2]{#2}%
  \newcommand*\fsize{\dimexpr\f@size pt\relax}%
  \newcommand*\lineheight[1]{\fontsize{\fsize}{#1\fsize}\selectfont}%
  \ifx\svgwidth\undefined%
    \setlength{\unitlength}{315.27048426bp}%
    \ifx\svgscale\undefined%
      \relax%
    \else%
      \setlength{\unitlength}{\unitlength * \real{\svgscale}}%
    \fi%
  \else%
    \setlength{\unitlength}{\svgwidth}%
  \fi%
  \global\let\svgwidth\undefined%
  \global\let\svgscale\undefined%
  \makeatother%
  \begin{picture}(1,1.01794047)%
    \lineheight{1}%
    \setlength\tabcolsep{0pt}%
    \put(0,0){\includegraphics[width=\unitlength,page=1]{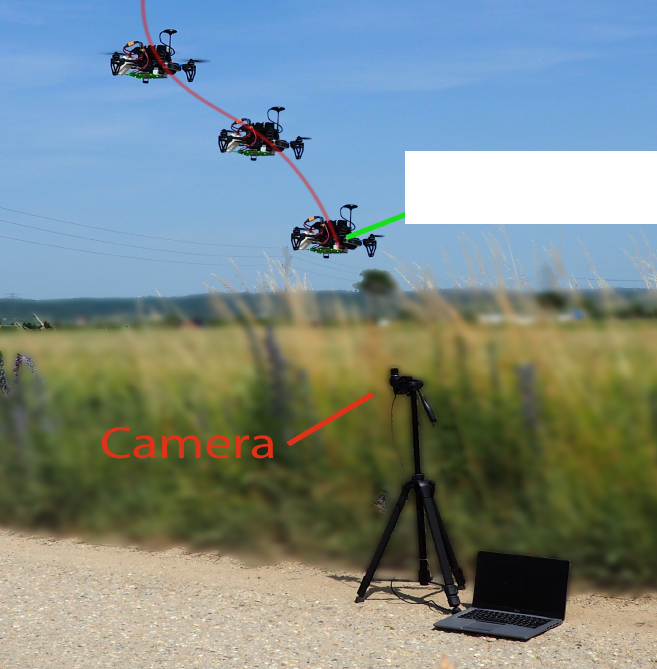}}%
    \put(0.64993564,0.6998279){\makebox(0,0)[lt]{\lineheight{1.25}\smash{\begin{tabular}[t]{l}{\footnotesize Marker}\end{tabular}}}}%
    \put(0,0){\includegraphics[width=\unitlength,page=2]{cam_on_ground.pdf}}%
    \put(0.09037732,0.29956152){\makebox(0,0)[lt]{\lineheight{1.25}\smash{\begin{tabular}[t]{l}{\footnotesize Camera}\end{tabular}}}}%
  \end{picture}%
\endgroup%

        \end{minipage}
        \hfill
        \begin{minipage}[t]{0.66\columnwidth}
            \vspace{0pt} 
            \centering
            \includegraphics[width=\linewidth]{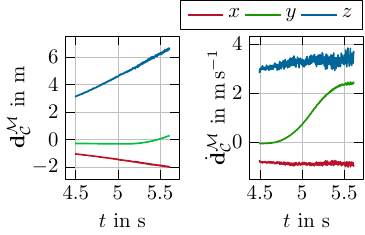} 
        \end{minipage}
        \caption{Camera on the ground.}
        \label{fig:cam_on_ground}
    \end{subfigure}
    \begin{subfigure}{\columnwidth}
        \begin{minipage}[t]{0.33\columnwidth}
            \vspace{5pt} 
            \centering
            \def\svgwidth{\textwidth}
\begingroup%
  \makeatletter%
  \providecommand\color[2][]{%
    \errmessage{(Inkscape) Color is used for the text in Inkscape, but the package 'color.sty' is not loaded}%
    \renewcommand\color[2][]{}%
  }%
  \providecommand\transparent[1]{%
    \errmessage{(Inkscape) Transparency is used (non-zero) for the text in Inkscape, but the package 'transparent.sty' is not loaded}%
    \renewcommand\transparent[1]{}%
  }%
  \providecommand\rotatebox[2]{#2}%
  \newcommand*\fsize{\dimexpr\f@size pt\relax}%
  \newcommand*\lineheight[1]{\fontsize{\fsize}{#1\fsize}\selectfont}%
  \ifx\svgwidth\undefined%
    \setlength{\unitlength}{315.27083029bp}%
    \ifx\svgscale\undefined%
      \relax%
    \else%
      \setlength{\unitlength}{\unitlength * \real{\svgscale}}%
    \fi%
  \else%
    \setlength{\unitlength}{\svgwidth}%
  \fi%
  \global\let\svgwidth\undefined%
  \global\let\svgscale\undefined%
  \makeatother%
  \begin{picture}(1,1.01795862)%
    \lineheight{1}%
    \setlength\tabcolsep{0pt}%
    \put(0,0){\includegraphics[width=\unitlength,page=1]{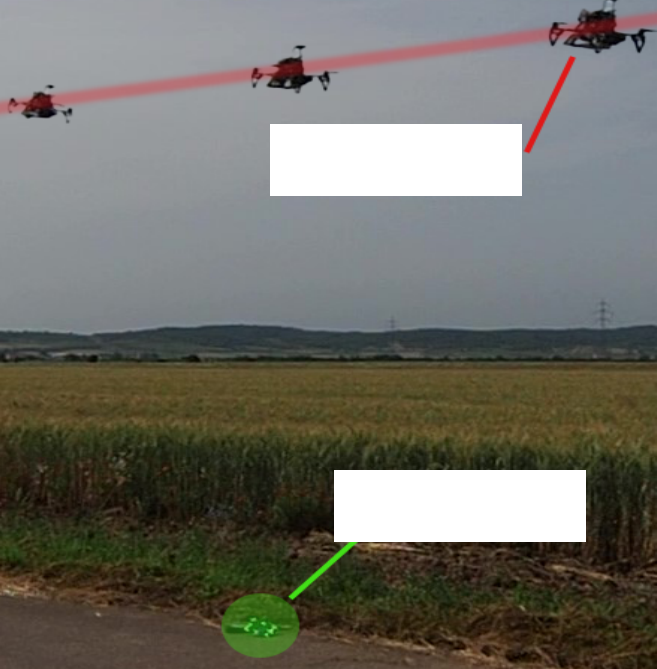}}%
    \put(0.45002532,0.74062025){\makebox(0,0)[lt]{\lineheight{1.25}\smash{\begin{tabular}[t]{l}{\footnotesize Camera}\end{tabular}}}}%
    \put(0.5421674,0.21481547){\makebox(0,0)[lt]{\lineheight{1.25}\smash{\begin{tabular}[t]{l}{\footnotesize Marker}\end{tabular}}}}%
  \end{picture}%
\endgroup%

        \end{minipage}
        \hfill
        \begin{minipage}[t]{0.66\columnwidth}
            \vspace{0pt} 
            \centering
            \includegraphics[width=\linewidth]{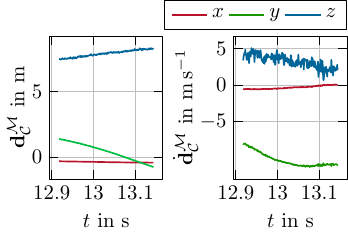} 
        \end{minipage}
        \caption{Camera on the drone.}
        \label{fig:cam_on_drone}
        \end{subfigure}
    
    \caption{The plot of the drone's trajectory and velocities during experiments.
    The derivative of the positional signal ${\vec{d}}_\mathcal{C}^\mathcal{M}$ was filtered with a moving average filter with a window length of 100 samples.
    }
    \label{fig:drone_2exp}
\end{figure}

Outdoor experiments were conducted to demonstrate the capability of the proposed system to detect and track motions at very high speeds. 
In the first scenario, the drone is equipped with an ALM and the event-based camera is mounted vertically on a tripod on the ground, see \cref{fig:cam_on_ground}. 
Although the drone moves at a maximum speed of \SI{4.5}{\meter\per\second} and ascends to a height of \SI{9}{\meter}, the proposed system is still able to capture the trajectories of the ALM, depicted in \cref{fig:cam_on_ground}. The position signals ${\vec{d}}_\mathcal{C}^\mathcal{M}$ indicate a low noise floor. The velocity signals $\dot{\vec{d}}_\mathcal{C}^\mathcal{M}$ are calculated based on the position signals with additional moving average filtering. 
Unlike in the first scenario, the camera is mounted on the drone and the ALM is static on the ground in the second scenario, as illustrated in \cref{fig:cam_on_drone}. The captured trajectories are shown in \cref{fig:cam_on_ground} when the drone is moving with an average speed of \SI{10}{\meter\per\second}. 
In both scenarios, the velocity in the $z$ direction is noisy due to the higher noise in the $z$ estimation by the PnP algorithm. 
Live videos of the two scenarios are provided in the supplementary material.

\label{sec:drone_ot_vs_slam}

\subsection{Limitations}
\textcolor{black}{
The performance of the proposed system is mainly limited by the resolution of the event-based camera and the power of the light source. 
Based on the sensor resolution, the ground sampling distance (GSD) using the \SI{8}{\milli \m} lens at \SI{10}{\m} is equal to \SI{0.61}{\centi\metre}. This sets an upper bound on the accuracy of the system, even with sub-pixel tracking precision and a precisely calibrated camera. Light intensity decreases with the square of the distance \cite{active_space}. Hence, LEDs with \SI{1}{\watt} at 10\% duty cycle were used in the experiments. Altogether, the maximum working distance is around 10m.
}
\textcolor{black}{
In favor of accuracy, ALMs with occluded LEDs are not tracked. However, occlusion of LEDs is detected to determine if the tracking is lost. An ALM is respawned when all LEDs are visible again. As the proposed system features a small marker size intended for outdoor usage, where other positioning systems are not applicable, LED occlusion is not a primary concern. 
}



\section{Conclusion}
\label{sec:conclusion}

This paper presents a fast and accurate vision-based localization system using an event-based camera with active LED markers. 
Our proposed method overcomes the limits of traditional marker-based localization systems, \ie low frame rate, motion blur, and high computational costs, by utilizing the advantages of an event-based camera. 
The proposed algorithm is simple but effective, achieving real-time performance with minimal latency below \SI{0.5}{\milli\second} and output rates above \SI{3}{\kilo \hertz} using a regular PC. 
The proposed tracking-based approach outperforms detection-based methods, especially in applications with very fast movements.
the position error normalized to the distance is constantly below \SI{1.87}{\percent} with a mean orientation error of \SI{0.738}{\degree}.
To the best of the authors' knowledge, the combination of the achieved precision at this output rate and latency was not achieved so far. For applications, where latency is not crucial, the output of the system can be filtered and fused with other modalities (IMU, RBG-based localization systems).
The proposed system can be used as a cheap relative localization/reference positioning system for outdoor applications as data collection where other systems can not be applied (dynamic scenes, fast motion). Also, the proposed method opens new possibilities for robotic applications where the high output rates and high precision of 6-DoF pose estimation are important, \eg dynamic handover tasks and pick-and-place tasks. 

{\small
\bibliographystyle{ieee_fullname}
\bibliography{bibliography}
}

\newpage
\clearpage
\appendix


\onecolumn
\begin{center}
    \section*{Appendix}    
\end{center}

\section{Kinematic Relations of the Experimental Setup}
\label{sec:kin_opt}

For the description of the kinematic relations of the experimental setup, the notation of a homogeneous transformation from the coordinate frame $\mathcal{A}$ to the coordinate frame $\mathcal{B}$
\begin{equation}
    \vec{H}_\mathcal{A}^\mathcal{B} = \begin{bmatrix}
        \vec{R}_\mathcal{A}^\mathcal{B} & \vec{d}_\mathcal{A}^\mathcal{B} \\
        0 & 1
    \end{bmatrix} \Comma
    \label{eq:homtransf}
\end{equation}
comprising the displacement $\vec{d}_\mathcal{A}^\mathcal{B} \in \mathbb{R}^3$ and the rotation matrix $\vec{R}_\mathcal{A}^\mathcal{B} \in \mathrm{SO}(3)$, 
is utilized. 

\begin{figure}{B}
	\centering
	\def\svgwidth{0.5\columnwidth}
	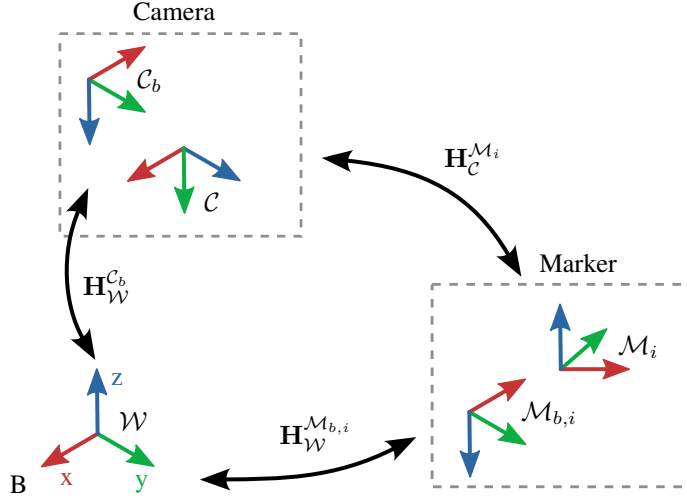
	\caption{Kinematic relations of the experimental setup. The coordinate system $\mathcal{W}$ indicates the base coordinate system of the \optitrack{} system with $\mathcal{C}_b$ and $\mathcal{M}_{b,i}$ designating the \optitrack{} markers of the camera base and the marker base $i$, respectively. The proposed method estimates the pose of the marker coordinate system $\mathcal{M}_i$ with respect to the optical center of the camera $\mathcal{C}$.} 
	\label{fig:kinematics_setup}
\end{figure}

\Cref{fig:kinematics_setup} gives an overview of the kinematic relations of the setup. 
Therein, the coordinate system $\mathcal{W}$ is the base frame of the \optitrack{} system, $\mathcal{C}_b$ denotes the frame of the \optitrack{} markers forming the rigid body of the event-based camera. 
Similarly, to detect the pose of the ALMs on the board, \optitrack{} markers are attached to each $i$-th ALM, forming the corresponding rigid body $\mathcal{M}_{b,i}$. 
The measurements of the \optitrack{} system yield the transformations $\vec{H}_\mathcal{W}^{\mathcal{C}_b}$  and $\vec{H}_\mathcal{W}^{\mathcal{M}_{b,i}}$ for the rigid body attached to the camera and the $i$-th ALM.
On the other hand, the ALM pose estimation using the event-based camera yields the transformation $\mathbf{H}_\mathcal{C}^{\mathcal{M}_i}$ of the $i$-th ALM with respect to the event-based camera's optical center $\mathcal{C}$. 

To fully determine the kinematic relations in \cref{fig:kinematics_setup} additional transformations, namely $\vec{H}_{\mathcal{C}_b}^\mathcal{C}$ and $\vec{H}_{\mathcal{M}_{b,i}}^{\mathcal{M}_i}$, are computed using optimization by fitting the closed kinematic chain to the ground truth. 
To this end, the pose of the $i$-th marker is expressed as 
\begin{equation}
    \bar{\vec{H}}_\mathcal{W}^{\mathcal{M}_i} = \vec{H}_\mathcal{W}^{\mathcal{C}_b}
    \hat{\vec{H}}_{\mathcal{C}_b}^\mathcal{C}
    \vec{H}_\mathcal{C}^{\mathcal{M}_i}\Comma
    \label{eq:h1}
\end{equation}
also equivalent to
\begin{equation}
    \bar{\bar{\vec{H}}}_\mathcal{W}^{\mathcal{M}_i} = \vec{H}_\mathcal{W}^{\mathcal{M}b,i} 
    \hat{\vec{H}}_{\mathcal{M}_{b,i}}^{\mathcal{M}_i} \FullStop
    \label{eq:h2}
\end{equation}

Using (\ref{eq:h1}) and (\ref{eq:h2}), the residual transformation reads as
\begin{equation}
    \tilde{\vec{H}}_{i} = \left( \bar{\bar{\vec{H}}}_\mathcal{W}^{\mathcal{M}_i} \right)^{-1}  \bar{\vec{H}}_\mathcal{W}^{\mathcal{M}_i} = 
     \begin{bmatrix}
        \tilde{\vec{R}}_i & \tilde{\vec{d}}_i \\
        0 & 1
    \end{bmatrix} 
    \FullStop
    \label{eq:hom_residual}
\end{equation}

With the residual pose (\ref{eq:hom_residual}), the hidden transformations $\{ \hat{\vec{H}}_{\mathcal{C}_b}^\mathcal{C}, \hat{\vec{H}}_{\mathcal{M}_{b,1}}^{\mathcal{M}_1}, \hat{\vec{H}}_{\mathcal{M}_{b,2}}^{\mathcal{M}_2} \} \in \mathrm{SE}(3)$ for the experimental setup comprising two markers $i \in \{1,2\}$ can be estimated over $k = 1, \dots, N$ measurements with the following optimization problem
\begin{equation}
    {\text{min}} \ \sum_{k=1}^{N} \sum_{i=1}^{2} \left[ \norm{\tilde{\vec{d}}_{i,k}}_2 + w \norm{\II - \tilde{\vec{R}}_{i,k}}_\text{F} \right]^2 \Comma
    \label{eq:opt_prob}
\end{equation}
where $\norm{\cdot}_\text{F}$ denotes the Frobenius norm.
The choice for the metric of the orientation error is motivated in \cite{Hu09} with an empirically chosen weighting value $w = \frac{1}{10}$. 
\section{Bias Adjustment Procedure}

\begin{figure}
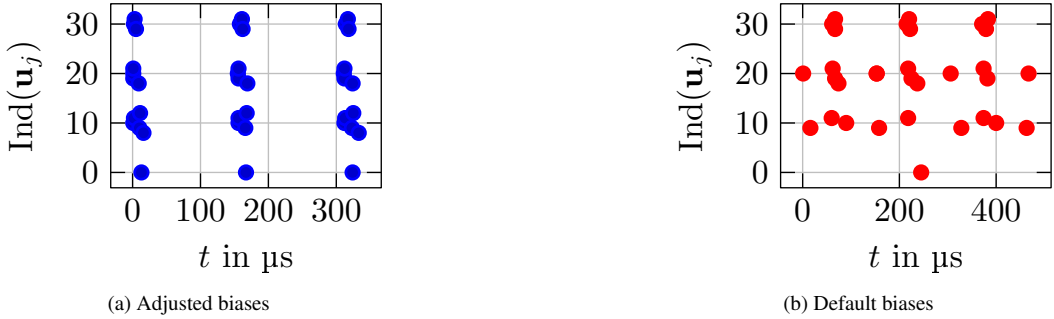

  \centering
  \begin{subfigure}[]{0.49\columnwidth}
    \centering
    \includegraphics[width=0.6\columnwidth]{graphics/bias/bias_good.pdf}
    \caption{Adjusted biases}
    \label{fig:bias_good}
  \end{subfigure}
  \hfill
  \begin{subfigure}[]{0.49\columnwidth}
    \centering
    \includegraphics[width=0.6\columnwidth]{graphics/bias/bias_bad.pdf}
    \caption{Default biases}
    \label{fig:bias_bad}
  \end{subfigure}
  \caption{Visualization of bias adjustment in an event camera using the IMX636ES sensor. The vertical axis represents the flattened indices of the pixels in the region of interest (ROI) around an LED light. The left plot demonstrates an optimal bias adjustment, where event fronts - closely packed clusters of events triggered by a rapid change in the scene (like a sudden LED blink) - are clearly distinct. No unwanted spurious events occur between these event fronts. The right plot, in contrast, displays the event distribution with the camera's default bias settings, showing a less distinct separation of event fronts. 
  } 
  \label{fig:bias}
\end{figure}

The IMX636ES sensor has five biases \cite{metavisionbiases}. The first two are the contrast sensitivity thresholds \emph{bias\_diff\_on} and \emph{bias\_diff\_off} that set the contrast thresholds for on and off events, respectively. The biases \emph{bias\_fo} and \emph{bias\_hpf} control the bandwidth by setting the illumination signal's low- and highpass filters. Lastly, \emph{bias\_refr} sets the refractory period, determining the duration for which the pixel is blind after each event. 
In the case of the IMX636ES sensor, the biases are expressed as relative offsets from the factory-trimmed default value, resulting in better portability of bias settings between different sensors. 

The goal during bias adjustment is to distinguish event fronts and reduce events caused by noise and dynamic changes in the scene. Event fronts are closely packed clusters of events, usually triggered by rapid, high-contrast changes in the scene, like the sudden blink of an LED light.
For illustration, refer to \cref{fig:bias} where the pixels of the ROI around a LED are plotted over time. The vertical axis shows the flattened indices $\text{Ind}(\vec{u}_j)$ of the pixels $\vec{u}_j$ in the ROI. 
\Cref{fig:bias_good} shows a good bias adjustment with clearly distinct event fronts.
In between event fronts, no unwanted spurious events occur. 

The selection of biases depends on the application. For active markers, the camera's default biases are an adverse choice, as depicted in \cref{fig:bias_bad}.
The event plots for the camera's default bias settings is depicted in \cref{fig:bias_bad}.

To adjust the biases, start with the default settings and prepare an online visualization as in \cref{fig:bias}. First, increase \emph{bias\_fo} until the event rate peaks. Gradually decrease the \emph{bias\_hpf} value until a significant drop in the event rate is observed. Increase the \emph{bias\_diff\_on} and \emph{bias\_diff\_off} values until the event rate drops significantly. To detect only single polarity events, increase either \emph{bias\_diff\_on} or \emph{bias\_diff\_off} to their maximum values. Adjust the \emph{bias\_refr} until the event fronts are as narrow as possible. As the bias settings affect the analog front end of the sensor, biases influence each other. For the fine adjustment, look at the visualization and the event rate and alter the settings until the best separation between event fronts is reached while keeping the event rate high. The biases used in the experiments are listed in \cref{tab:biases}.

\begin{table}[htbp]
    \centering
    \caption{List of biases for pose estimation with active LED markers used in the experiments.}
    \label{tab:biases}
    \begin{tabular}{l r}
        \hline
        \textbf{Parameter} & \textbf{Value} \\
        \hline
        \emph{bias\_diff\_off} & 177 \\
        \emph{bias\_diff\_on} & 0 \\
        \emph{bias\_fo} & 10 \\
        \emph{bias\_hpf} & 120 \\
        \emph{bias\_refr} & 0 \\
        \hline
    \end{tabular}
\end{table}

\section{Additional experiment on 6-DoF position estimation for a quadcopter}
\begin{figure}[h]
    \centering
    \includegraphics[width = 0.9\columnwidth]{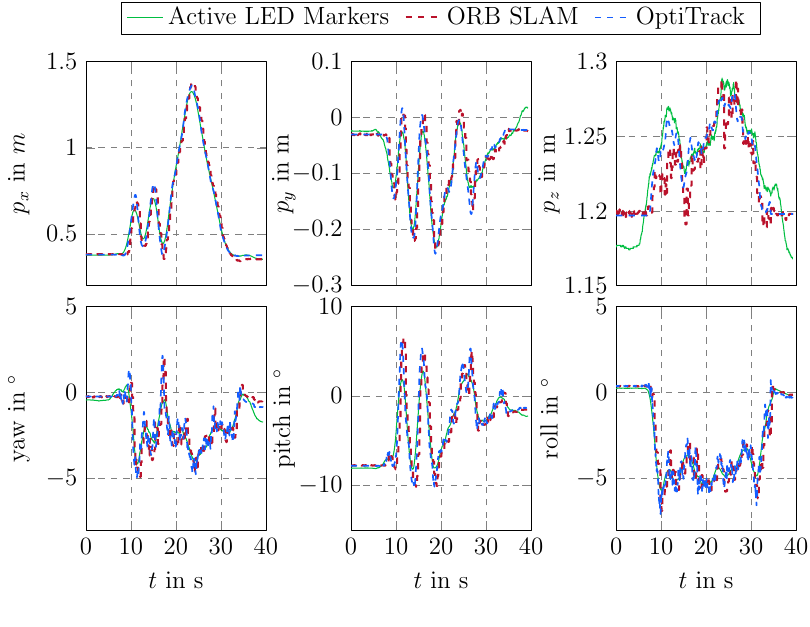}
    \caption{Experiment 2: The estimated poses from the proposed method, the ORB-SLAM2, and the ground truths from \optitrack{} are illustrated in green, red, and blue, respectively.}%
    \label{fig: trial 1}%
\end{figure}

\begin{figure}[h]
    \centering
    \includegraphics[width = 0.9\columnwidth]{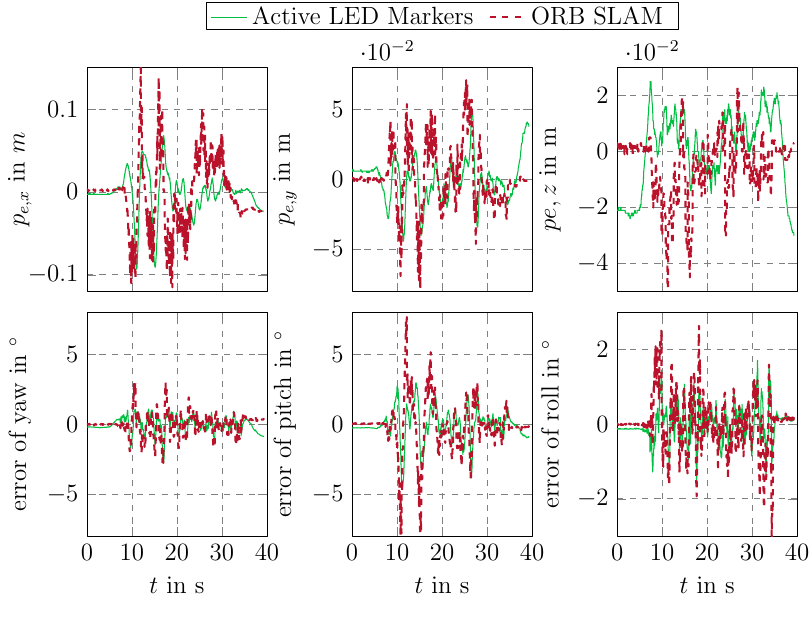}
    \caption{Experiment 2: The error plots of the corresponding estimates, depicted in Fig. \ref{fig: trial 1} with respect to the ground-truth measurements from the \optitrack{}.}%
    \label{fig: trial 1 error}%
\end{figure}

In the second experiment, the quadcopter is slowly operated in a space of \SI{1.2}{\meter} in the $x$-direction, \SI{0.3}{\meter} in  $y$-direction, and \SI{0.1}{\meter} in  $z$-direction for about \SI{40}{\second}. 
The position estimates and the resulting errors for this case are depicted in Fig. \ref{fig: trial 1} and Fig. \ref{fig: trial 1 error}. The translational errors obtained from the proposed method, \ie, $p_{e,x} \in [-0.06,0.03]\:\SI{}{\meter}$, $p_{e,y} \in [-0.02,-0.04]\:\SI{}{\meter}$, and $p_{e,z} \in [-0.02,0.02]\:\SI{}{\meter}$, are smaller than the ones obtained from the ORB-SLAM2,  \ie, $p_{e,x} \in [-0.1,0.1]\:\SI{}{\meter}$, $p_{e,y} \in [-0.06,-0.05]\:\SI{}{\meter}$, $p_{e,z}\:\in [-0.04,-0.02] \SI{}{\meter}$. On the other hand, the rotational errors obtained from the two methods are nearly similar. However, larger measurement spikes are observed with the ORB-SLAM2 method. 



\end{document}